\documentclass[sigconf,screen]{acmart}

\AtBeginDocument{%
  \providecommand\BibTeX{{%
    \normalfont B\kern-0.5em{\scshape i\kern-0.25em b}\kern-0.8em\TeX}}}

\acmConference[WSDM '24]{The 17th ACM International Conference on Web Search and Data Mining}{March 4--8, 2024}{Mérida, Yucatán, Mexico}
\usepackage{multirow} 
\usepackage{enumitem}
\usepackage[htt]{hyphenat}
\begin{document}

\title[Leveraging Large Language Models for Conversational Multi-Doc QA]{The First Place Solution of WSDM Cup 2024: Leveraging Large Language Models for Conversational Multi-Doc QA}


\author{Yiming Li}
\authornote{Equal contribution.}
\affiliation{%
  \institution{Beijing Key Laboratory of Mobile Computing and Pervasive Device, Institute of Computing Technology, Chinese Academy of Sciences}
  \institution{University of Chinese Academy of Sciences}
  \city{Beijing}
  \country{China}}
\email{liyiming22s1@ict.ac.cn}

\author{Zhao Zhang}
\authornotemark[1]
\affiliation{%
  \institution{CAS Key Lab of Network Data Science and Technology, Institute of Computing Technology, Chinese Academy of Sciences}
  \institution{University of Chinese Academy of Sciences}
  \city{Beijing}
  \country{China}
}
\email{zhangzhao22s@ict.ac.cn}


\begin{abstract}
Conversational multi-doc question answering aims to answer specific questions based on the retrieved documents as well as the contextual conversations. In this paper, we introduce our winning approach for the “Conversational Multi-Doc QA” challenge in WSDM Cup 2024, which exploits the superior natural language understanding and generation capability of Large Language Models (LLMs). We first adapt LLMs to the task, then devise a hybrid training strategy to make the most of in-domain unlabeled data. Moreover, an advanced text embedding model is adopted to filter out potentially irrelevant documents and several approaches are designed and compared for the model ensemble. Equipped with all these techniques, our solution finally ranked 1st place in WSDM Cup 2024, surpassing its rivals to a large extent. The source codes have been released at \url{https://github.com/zhangzhao219/WSDM-Cup-2024}.
\end{abstract}

\vspace{-1.0em}




\maketitle{}

\section{Introduction}
Conversational question answering \citep{10.1007/s10115-022-01744-y}, which aims to generate correct and meaningful answers according to users' intents identified from the dialog, plays a crucial role in modern search engines and conversational systems. However, it is still challenging, especially with current or trending topics as timely knowledge is inaccessible during the training stage of the language model. Although providing multiple relevant documents as contextual information seems feasible, the model is still at risk of being overwhelmed or misled by the massive input. Based on the real-world textual data, Xiaohongshu, the WSDM Cup 2024\footnote{\url{https://sites.google.com/view/wsdm24-docqa}} presents a challenge of “Conversational Multi-Doc QA” to encourage further exploration of the problem.

Recently, LLMs, such as ChatGPT, have demonstrated impressive performance on several natural language processing tasks. It is promising to solve the challenge by leveraging the understanding and reasoning abilities of LLMs. However, lots of factors, including the design of training configurations and the existence of irrelevant documents, still hinder the improvement of generation quality. 

In this work, to unleash the power of LLMs, we first formulate the task as a multi-turn conditional generation problem with different LLMs. Then, a multi-stage hybrid training pipeline is conducted to incorporate an unlabeled eval set as an additional training corpus. To remove potentially irrelevant information, we implement certain strategies, including a state-of-the-art embedding model namely Nomic Embed \citep{nussbaum2024nomic} to compute the similarity score between inputs and documents. Finally, several methods are considered to approximately evaluate the quality of answers generated by various LLMs before selecting the best response as the final answer for the model ensemble. Experimental results show that our solution achieves the highest score on each evaluation metric, far beyond the teams behind us, while ablation studies also suggest the effectiveness of proposed techniques. 

\section{Preliminary}
\subsection{Dataset} 
For the competition, participants are required to train a model to produce answers corresponding to dialogue history (composed of sequential qa pairs), reference documents, and the final question. To this end, each training sample is organized as follows,

\texttt{history: \textbf{\{q1\}\{a1\}\{q2\}\{a2\}...\{qn\}\{an\}}}

\texttt{documents: \textbf{\{d1\}\{d2\}...\{dn\}}}

\texttt{question: \textbf{\{q\}}}

\texttt{answer: \textbf{\{a\}}}

\noindent {But for the eval and test set, the answer field is invisible while an additional field, namely keywords is incorporated to comprehensively evaluate the generated answer, which will be detailed later.} 
\vspace{-1.5em}

\subsection{Evaluation} 

Three metrics are involved to evaluate the lexical and semantic relatedness of generated answers. The definitions are detailed as follows,
\begin{itemize}[labelsep = .5em, leftmargin = 5pt, itemindent = 0.5em]
\item {\bf ROUGE-L} \quad The intuition is that the longer the LCS (the longest common subsequence) of two sentences is, the more similar the two sentences are \citep{lin-2004-rouge}. As a result, LCS-based F-measure can be used to estimate the similarity between two
sentences $X$ of length $m$ and $Y$ of length $n$, 
$$R_{lcs} = \frac{LCS(X,Y)}{m}$$
$$P_{lcs} = \frac{LCS(X,Y)}{n}$$
$$F_{lcs} = \frac{(1+\beta^2)R_{lcs}P_{lcs}}{R_{lcs}+\beta^2P_{lcs}}$$

Specifically, word-level ROUGE-L (abbreviated as W-ROUGE-L below) calculates $F_{lcs}$ of the words in sentences, which is more focused on the accuracy of specific words, while character-level ROUGE-L (abbreviated as C-ROUGE-L below) pays more attention to word forms, grammar, and punctuation because of calculating $F_{lcs}$ of the characters.

\item {\bf Keywords Recall} \quad Keywords Recall (abbreviated as KR below) focuses on whether the specific keywords of truth sentence $X$ have appeared in the generated answer $Y$. If there are $m$ reference keywords in $X$, and $n$ of them appear in $Y$, 
$$KR=\frac{n}{m}$$
\end{itemize}

\section{Methodology}

\subsection{Generation Baseline with LLMs} 
To adapt LLM to this task, we carefully design the input format and concatenate each textual component together in the following order,

$u$ = \texttt{\textbf{\{q1\}\{a1\}\{q2\}\{a2\}...\{qn\}\{an\}\{q\}\{d1\}\{d2\}...\{dn\}\{a\}}}

\noindent {Note that we exclude special tokens (e.g., <s>, [INST]) in above line for simplicity.}

Then, the model $\theta$ can be trained by maximizing the log-likelihood over the whole sequence.
\vspace{-0.25em}
$$\mathcal{L}_{gen}=-\sum_{i=1}^{|\boldsymbol{u}|} m_i \log{p(u_{i}|, \boldsymbol{u}_{<i};\theta)}$$
where $p(u_{i}|, \boldsymbol{u}_{<i};\theta)$ is the probability to select a token $u_{i}$ at step $i$ given previous tokens $\boldsymbol{u}_{<i}$, and $m_i$ is the loss mask for the $i$th token. Specifically, there are two modes to determine $m_i$: 1) the single-turn mode, which means that $m_i = 1$ if and only if $u_i$ belongs to \texttt{\textbf{\{a\}}}. 2) the multi-turn mode, $m_i = 1$ as long as $u_i$ belongs to \texttt{\textbf{\{a\}}} or \texttt{\textbf{\{ai\}}}. We conduct a toy experiment to examine them with the Llama2-13B-base model \citep{touvron2023llama}, the results are shown in Table \ref{turn_num}. It can be seen that the multi-turn mode results in a better performance as it forces the LLM to pay more attention to contextual information.

\begin{table}[ht]
\centering
\caption{\label{turn_num}
Performance of the specific model trained with single-turn and multi-turn modes on the eval dataset.
}
\vspace{-1.0em}
\begin{tabular}{ccc}
\hline
\textbf{Mode} & \textbf{C-ROUGE-L} & \textbf{KR} \\
\hline
single-turn & $\text{0.5761}$ & $\text{0.6291}$ \\
multi-turn & $\textbf{0.5845}$ & $\textbf{0.6359}$ \\
\hline
\end{tabular}
\end{table}

After deciding on the input format and mask mode, we compare lots of off-the-shelf LLMs, which are either pretrained only or instruction tuned. As shown in Table \ref{different_LLM}, the SOLAR-10.7B-Instruct model surpasses its counterparts a lot on the eval dataset, which uses depth up-scaling to scale LLM and fine-tuned for instruction-following capabilities \citep{kim2023solar}. Therefore, it is chosen as our backbone in the subsequent experiments. 

\begin{table}[ht]
\centering
\caption{\label{different_LLM}
Performance of different LLMs on the eval dataset.
}
\vspace{-1.0em}
\begin{tabular}{lcc}
\hline
\textbf{LLM} & \textbf{C-ROUGE-L} & \textbf{KR} \\
\hline
ChatGPT (zero-shot) \citep{chatgpt} & $\text{0.4815}$ & $\text{0.5387}$ \\
GPT-4 (zero-shot) \citep{openai2023gpt4} & $\text{0.5069}$ & $\text{0.5537}$ \\
Yi-6B \citep{Yi} & $\text{0.5286}$ & $\text{0.6464}$ \\
ChatGLM3-6B \citep{zeng2022glm} & $\text{0.5740}$ & $\text{0.6068}$ \\
ChatGLM3-6B-Base \citep{zeng2022glm} & $\text{0.5782}$ & $\text{0.6166}$ \\
DeepSeek 7B Base \citep{deepseek-llm} & $\text{0.5783}$ & $\text{0.6184}$ \\
Llama 2 Chat 13B \citep{touvron2023llama} & $\text{0.5821}$ & $\text{0.6266}$ \\
Yi-6B-Chat \citep{Yi} & $\text{0.5833}$ & $\text{0.6365}$ \\
Llama 2 13B \citep{touvron2023llama} & $\text{0.5845}$ & $\text{0.6359}$ \\
Mistral 7B \citep{jiang2023mistral} & $\text{0.6031}$ & $\text{0.6489}$ \\
Mistral 7B-Instruct \citep{jiang2023mistral} & $\text{0.6048}$ & $\text{0.6558}$ \\
SOLAR 10.7B \citep{kim2023solar} & $\text{0.6099}$ & $\text{0.6627}$ \\
SOLAR 10.7B-Instruct \citep{kim2023solar} & $\textbf{0.6104}$ & $\textbf{0.6691}$ \\
\hline
\end{tabular}
\end{table}

\subsection{Hybrid Training}
Appropriate labeled texts from a similar distribution may contribute a lot to the improvement of LLMs' generation performance. During phase 2, we propose to utilize a well-trained model to produce (pseudo) answers for the eval dataset before adding them to the original training set to finetune a new model from scratch. The intuition for the above hybrid training strategy is two-fold, on one hand, it can be viewed as the knowledge distillation process on in-domain unlabeled data, on the other hand, since we only generate the final target \texttt{\textbf{\{a\}}} in a pseudo labeling manner, \texttt{\textbf{\{ai\}}} are still officially annotated, which may be beneficial for the multi-turn setting. Note that we do not further involve the test dataset for hybrid training as it may over-calibrate the model and thus weaken the model performance in the final evaluation, which is also examined by our empirical practices. 

\subsection{Noisy Document Filter} 

\begin{figure}[ht]
\centering
\includegraphics[scale=0.625]{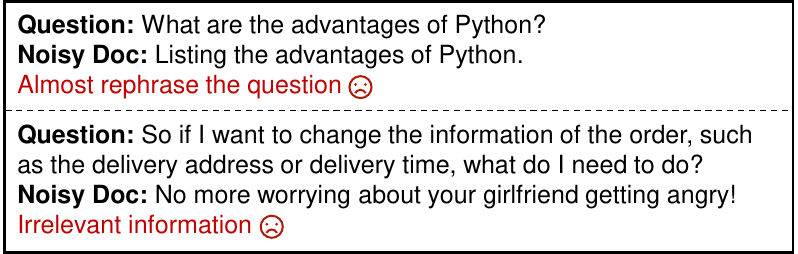}
\caption{Two examples of noisy documents}
\vspace{-1.0em}
\label{fig:doc_example}
\end{figure}

There is no doubt that high-quality reference documents can not only help mitigate the hallucination phenomena but also enhance the inference quality of LLMs \citep{li-etal-2023-large}. After manually scrolling through the whole dataset, we find there are mainly two types of noisy documents, the examples of which can be found in Figure \ref{fig:doc_example}.
\begin{itemize}[labelsep = .5em, leftmargin = 5pt, itemindent = 0.5em]
\item Documents almost rephrasing the question, which shares extremely high relevant scores with the document.

\item Documents full of irrelevant information, so they share extremely low relevant scores with the question or history.
\end{itemize}

Therefore, it is vital to quantify the relevance without the existence of ground truth answers. From both semantic and lexical views, we come up with the following two indicators, 

\begin{itemize}[labelsep = .5em, leftmargin = 5pt, itemindent = 0.5em]
\item {\bf Embedding-level Cosine Similarity}\quad
We adopt an advanced text embedding model, Nomic Embed, to compute the cosine similarity between documents and the corresponding question (or together with conversational history).

\item {\bf Word- or character-level ROUGE-L}\quad As illustrated before, the ROUGE-L scores can be viewed as lexical relatedness criteria.

\end{itemize}

Practically, we set a higher threshold $\tau_h$ and a lower one $\tau_l$ for each indicator separately, before screening out reference documents whose corresponding scores $\geq \tau_h$ or $\leq \tau_l$ for manual inspection. As a result, we filter out 193 noisy documents in phase 2.

Moreover, prior work \citep{liu2023lost} suggests that important passages placed at the beginning or end of the input can be better comprehended by LLMs. However, we find that there is a strong correlation between document indexes and their relative orders occurring at the officially annotated answers, which means that reranking the reference documents can result in severe performance degradation.

\subsection{Model Ensemble}
The model ensemble has proven to be effective in discriminative tasks, however, it is rarely explored under generative settings. In this work, we propose to approximately evaluate the quality of generated answers from different models and then select the best one as the final result. Suppose that given a test sample, we have $M$ candidate responses to aggregate, for each candidate $r_i$, we calculate the relevance scores $s(r_i, r_j)$ between $r_i$ and $r_j (j = 1, \cdots, M, j\neq i)$ and add them together as the quality score $q_i$ for $r_i$ ($q_i = \sum_j s(r_i, r_j)$). Similarly, the relevance quantizers can be embedding-level cosine similarity (denoted as emb\_a\_s), word-level ROUGE-L (denoted as word\_a\_f), and character-level ROUGE-L (denoted as char\_a\_f). The motivation is that the final answer should be a representative who reaches a consensus with the most candidates.

\section{Experiments}

\subsection{Experimental Settings}

Thanks to the LLM fine-tuning framework provided by modelscope\footnote{\url{https://github.com/modelscope/swift}}, we can fine-tune and infer our model easily. All our experiments are carried out on NVIDIA A100 80G and V100 32G GPUs. As for the fine-tuning process, we employ configuration for the LoRA \citep{hu2021lora} fine-tuning, defined with the following parameters: lora\_rank (rank of the low-rank matrices) set to 8, lora\_alpha (scaling factor for learning rate) at 16, lora\_dropout to manage overfitting set at 0.05, and lora\_target\_modules focused on all modules. The maximum length of input sequences was set to 3072. The AdamW \citep{loshchilov2019decoupled} optimizer is employed for training with a learning rate set to 1e-4 with a warm-up proportion of 0.03 and a batch size of 1 on each GPU. This model is trained for 4 epochs via deepspeed ZeRO-2\footnote{\url{https://www.microsoft.com/en-us/research/blog/ZeRO-2-deepspeed-shattering-barriers-of-deep-learning-speed-scale/}}, and we choose checkpoint 1700 as our best model. As for the inferring process, we set do\_sample=false to use greedy decoding and ensure stable output. The repetition penalty is adjusted to be between 1.00 and 1.02 and max\_new\_tokens is set to 512. We use vLLM \citep{kwon2023efficient} to accelerate our inference process, and it takes about 40 minutes to infer our final result on a V100 32G GPU.

\subsection{Competition Results}

Table \ref{final_result} lists the final results of this competition. As shown, our solution, which aggregates outcomes from 8 different models, achieves 1.6\%, 0.9\%, and 2.3\% absolute performance gains on W-ROUGE-L, C-ROUGE-L, and KR respectively when compared to the 2nd place. Besides, it is noteworthy that our single model can also yield better performance than others, suggesting the effectiveness of our pipelines.

\begin{table}[ht]
\small
\centering
\caption{\label{final_result}
Top scores of the competition. We also list the results of our single model (marked with * ) for comparison.
}
\vspace{-1.0em}
\begin{tabular}{cccccc}
\hline
\textbf{Rank} & \textbf{Participant} & \textbf{W-ROUGE-L} & \textbf{C-ROUGE-L} & \textbf{KR}\\
\hline
\text{1} & \text{regtrh (ours)} & $\textbf{0.46536}$ & $\textbf{0.62084}$ & $\textbf{0.69535}$ \\
\text{-} & \text{regtrh* (ours)} & $\text{0.45548}$ & $\text{0.61420}$ & $\text{0.68359}$ \\
\text{2} & \text{wangkxu} & $\text{0.44961}$ & $\text{0.61183}$ & $\text{0.67230}$ \\
\text{3} & \text{zhangmin186} & $\text{0.45098}$ & $\text{0.61048}$ & $\text{0.66239}$ \\
\text{4} & \text{Ted} & $\text{0.45013}$ & $\text{0.61030}$ & $\text{0.66513}$ \\
\text{5} & \text{tilbur} & $\text{0.44554}$ & $\text{0.60489}$ & $\text{0.67718}$ \\

\hline
\end{tabular}
\end{table}

\subsection{Ablation Studies}

\textbf{Ablation Study of the Noisy Document Filter.}\quad Table \ref{ndf} shows the experimental results of our single model inferred with and without the noisy document filter. We find that it marginally improves the final scores since the provided documents are carefully chosen by the cup organizers and LLMs can somewhat tell apart the underlying distractors.

\begin{table}[ht]
\small
\centering
\caption{\label{ndf}
Ablation study of the noisy document filter on the test dataset.
}
\vspace{-1.0em}
\begin{tabular}{cccc}
\hline
\textbf{Noisy Document Filter} & \textbf{W-ROUGE-L} & \textbf{C-ROUGE-L} & \textbf{KR}\\
\hline
- & $\text{0.45547}$ & $\text{0.61368}$ & $\text{0.68243}$ \\
\checkmark & $\textbf{0.45548}$ & $\textbf{0.61420}$ & $\textbf{0.68359}$ \\

\hline
\end{tabular}
\end{table}

\noindent{\textbf{Ablation Study of the Hybrid Training Strategy.}}\quad We verify the effects of the proposed hybrid training strategy in Table \ref{hybrid_training}. As seen, incorporating the eval set with the corresponding pseudo targets can largely boost the generation quality, especially for the keywords recall score. But further incorporation of the test set has little effect, which validates our design choice.

\begin{table}[ht]
\small
\centering
\caption{\label{hybrid_training}
Ablation study of the hybrid training strategy on the test dataset.
}
\vspace{-1.0em}
\begin{tabular}{cccc}
\hline
\textbf{Method} & \textbf{W-ROUGE-L} & \textbf{C-ROUGE-L} & \textbf{KR}\\
\hline
\text{-} & $\textbf{0.45547}$ & $\textbf{0.61368}$ & $\textbf{0.68243}$ \\
\text{- hybrid training (eval set)} & $\text{0.45284}$ & $\text{0.61308}$ & $\text{0.67154}$ \\
\text{+ hybrid training (test set)} & $\text{0.45476}$ & $\text{0.61343}$ & $\text{0.67626}$ \\

\hline
\end{tabular}
\end{table}

\begin{figure}[ht]
\centering
\includegraphics[scale=0.275]{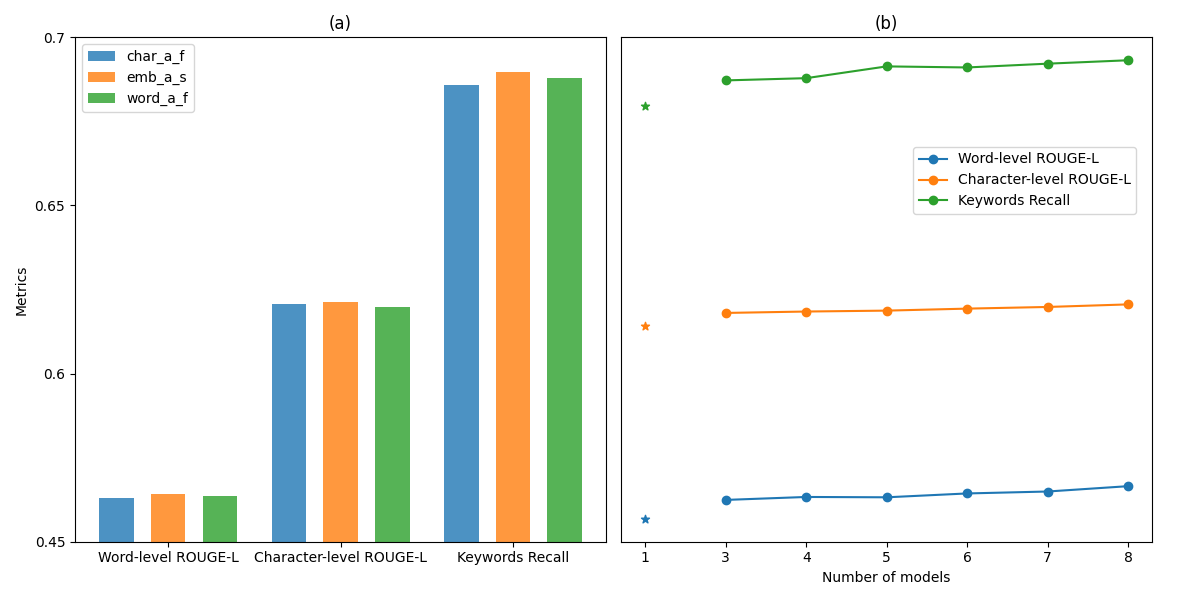}
\vspace{-1.0em}
\caption{(a) generation performances on the test dataset with different ensemble approaches (5 candidates for ensemble); (b) generation performances on the test dataset with different numbers of candidates for ensemble (with the approach working best in (a));}
\label{fig:ensemble}
\end{figure}

\noindent{\textbf{Ablation Study of the Model Ensemble.}}\quad We first compare different ensemble methods as displayed in Figure \ref{fig:ensemble} (a). Although the mentioned methods are all competitive on ROUGE scores, emb\_a\_s brings about many more improvements in keywords recall, thereby being selected as our final ensemble method. Then, a parameter analysis of the number of candidates for the ensemble is conducted. As seen from Figure \ref{fig:ensemble} (b), more candidates generally lead to better performance. Due to the limited time and budget, we finally defined the number as 8.

\vspace{-1.0em}
\section{Conclusion}

In this paper, we detail our winning solution to the task of "Conversational Multi-Doc QA" in WSDM Cup 2024. Leveraging the abilities of LLMs, we use the SOLAR-10.7B-Instruct model as the backbone, with the combination of hybrid training, noisy document filter, and selecting the best response by evaluating the quality from 8 results as our final submission. Our solution won 1st place on the public leaderboard.

\section*{Acknowledgments}

We thank everyone who offers advice to us and everyone associated with organizing and sponsoring the WSDM Cup 2024. We also express our gratitude to the organizations that provided us with computing resources.

\bibliographystyle{ACM-Reference-Format}
\bibliography{sample-base}


\begin{thebibliography}{16}


\ifx \showCODEN    \undefined \def \showCODEN     #1{\unskip}     \fi
\ifx \showDOI      \undefined \def \showDOI       #1{#1}\fi
\ifx \showISBNx    \undefined \def \showISBNx     #1{\unskip}     \fi
\ifx \showISBNxiii \undefined \def \showISBNxiii  #1{\unskip}     \fi
\ifx \showISSN     \undefined \def \showISSN      #1{\unskip}     \fi
\ifx \showLCCN     \undefined \def \showLCCN      #1{\unskip}     \fi
\ifx \shownote     \undefined \def \shownote      #1{#1}          \fi
\ifx \showarticletitle \undefined \def \showarticletitle #1{#1}   \fi
\ifx \showURL      \undefined \def \showURL       {\relax}        \fi
\providecommand\bibfield[2]{#2}
\providecommand\bibinfo[2]{#2}
\providecommand\natexlab[1]{#1}
\providecommand\showeprint[2][]{arXiv:#2}

\bibitem[DeepSeek-AI(2024)]%
        {deepseek-llm}
\bibfield{author}{\bibinfo{person}{DeepSeek-AI}.} \bibinfo{year}{2024}\natexlab{}.
\newblock \bibinfo{title}{DeepSeek LLM: Scaling Open-Source Language Models with Longtermism}.
\newblock
\newblock
\showeprint[arxiv]{2401.02954}~[cs.CL]
\urldef\tempurl%
\url{https://github.com/deepseek-ai/DeepSeek-LLM}
\showURL{%
\tempurl}


\bibitem[Hu et~al\mbox{.}(2021)]%
        {hu2021lora}
\bibfield{author}{\bibinfo{person}{Edward~J. Hu}, \bibinfo{person}{Yelong Shen}, \bibinfo{person}{Phillip Wallis}, \bibinfo{person}{Zeyuan Allen-Zhu}, \bibinfo{person}{Yuanzhi Li}, \bibinfo{person}{Shean Wang}, \bibinfo{person}{Lu Wang}, {and} \bibinfo{person}{Weizhu Chen}.} \bibinfo{year}{2021}\natexlab{}.
\newblock \bibinfo{title}{LoRA: Low-Rank Adaptation of Large Language Models}.
\newblock
\newblock
\showeprint[arxiv]{2106.09685}~[cs.CL]


\bibitem[Jiang et~al\mbox{.}(2023)]%
        {jiang2023mistral}
\bibfield{author}{\bibinfo{person}{Albert~Q. Jiang}, \bibinfo{person}{Alexandre Sablayrolles}, \bibinfo{person}{Arthur Mensch}, \bibinfo{person}{Chris Bamford}, \bibinfo{person}{Devendra~Singh Chaplot}, \bibinfo{person}{Diego de~las Casas}, \bibinfo{person}{Florian Bressand}, \bibinfo{person}{Gianna Lengyel}, \bibinfo{person}{Guillaume Lample}, \bibinfo{person}{Lucile Saulnier}, \bibinfo{person}{Lélio~Renard Lavaud}, \bibinfo{person}{Marie-Anne Lachaux}, \bibinfo{person}{Pierre Stock}, \bibinfo{person}{Teven~Le Scao}, \bibinfo{person}{Thibaut Lavril}, \bibinfo{person}{Thomas Wang}, \bibinfo{person}{Timothée Lacroix}, {and} \bibinfo{person}{William~El Sayed}.} \bibinfo{year}{2023}\natexlab{}.
\newblock \bibinfo{title}{Mistral 7B}.
\newblock
\newblock
\showeprint[arxiv]{2310.06825}~[cs.CL]


\bibitem[Kim et~al\mbox{.}(2023)]%
        {kim2023solar}
\bibfield{author}{\bibinfo{person}{Dahyun Kim}, \bibinfo{person}{Chanjun Park}, \bibinfo{person}{Sanghoon Kim}, \bibinfo{person}{Wonsung Lee}, \bibinfo{person}{Wonho Song}, \bibinfo{person}{Yunsu Kim}, \bibinfo{person}{Hyeonwoo Kim}, \bibinfo{person}{Yungi Kim}, \bibinfo{person}{Hyeonju Lee}, \bibinfo{person}{Jihoo Kim}, \bibinfo{person}{Changbae Ahn}, \bibinfo{person}{Seonghoon Yang}, \bibinfo{person}{Sukyung Lee}, \bibinfo{person}{Hyunbyung Park}, \bibinfo{person}{Gyoungjin Gim}, \bibinfo{person}{Mikyoung Cha}, \bibinfo{person}{Hwalsuk Lee}, {and} \bibinfo{person}{Sunghun Kim}.} \bibinfo{year}{2023}\natexlab{}.
\newblock \bibinfo{title}{SOLAR 10.7B: Scaling Large Language Models with Simple yet Effective Depth Up-Scaling}.
\newblock
\newblock
\showeprint[arxiv]{2312.15166}~[cs.CL]


\bibitem[Kwon et~al\mbox{.}(2023)]%
        {kwon2023efficient}
\bibfield{author}{\bibinfo{person}{Woosuk Kwon}, \bibinfo{person}{Zhuohan Li}, \bibinfo{person}{Siyuan Zhuang}, \bibinfo{person}{Ying Sheng}, \bibinfo{person}{Lianmin Zheng}, \bibinfo{person}{Cody~Hao Yu}, \bibinfo{person}{Joseph Gonzalez}, \bibinfo{person}{Hao Zhang}, {and} \bibinfo{person}{Ion Stoica}.} \bibinfo{year}{2023}\natexlab{}.
\newblock \showarticletitle{Efficient Memory Management for Large Language Model Serving with PagedAttention}. In \bibinfo{booktitle}{\emph{Proceedings of the 29th Symposium on Operating Systems Principles}} (<conf-loc>, <city>Koblenz</city>, <country>Germany</country>, </conf-loc>) \emph{(\bibinfo{series}{SOSP '23})}. \bibinfo{publisher}{Association for Computing Machinery}, \bibinfo{address}{New York, NY, USA}, \bibinfo{pages}{611–626}.
\newblock
\showISBNx{9798400702297}
\urldef\tempurl%
\url{https://doi.org/10.1145/3600006.3613165}
\showDOI{\tempurl}


\bibitem[Li et~al\mbox{.}(2023)]%
        {li-etal-2023-large}
\bibfield{author}{\bibinfo{person}{Daliang Li}, \bibinfo{person}{Ankit~Singh Rawat}, \bibinfo{person}{Manzil Zaheer}, \bibinfo{person}{Xin Wang}, \bibinfo{person}{Michal Lukasik}, \bibinfo{person}{Andreas Veit}, \bibinfo{person}{Felix Yu}, {and} \bibinfo{person}{Sanjiv Kumar}.} \bibinfo{year}{2023}\natexlab{}.
\newblock \showarticletitle{Large Language Models with Controllable Working Memory}. In \bibinfo{booktitle}{\emph{Findings of the Association for Computational Linguistics: ACL 2023}}, \bibfield{editor}{\bibinfo{person}{Anna Rogers}, \bibinfo{person}{Jordan Boyd-Graber}, {and} \bibinfo{person}{Naoaki Okazaki}} (Eds.). \bibinfo{publisher}{Association for Computational Linguistics}, \bibinfo{address}{Toronto, Canada}, \bibinfo{pages}{1774--1793}.
\newblock
\urldef\tempurl%
\url{https://doi.org/10.18653/v1/2023.findings-acl.112}
\showDOI{\tempurl}


\bibitem[Lin(2004)]%
        {lin-2004-rouge}
\bibfield{author}{\bibinfo{person}{Chin-Yew Lin}.} \bibinfo{year}{2004}\natexlab{}.
\newblock \showarticletitle{{ROUGE}: A Package for Automatic Evaluation of Summaries}. In \bibinfo{booktitle}{\emph{Text Summarization Branches Out}}. \bibinfo{publisher}{Association for Computational Linguistics}, \bibinfo{address}{Barcelona, Spain}, \bibinfo{pages}{74--81}.
\newblock
\urldef\tempurl%
\url{https://aclanthology.org/W04-1013}
\showURL{%
\tempurl}


\bibitem[Liu et~al\mbox{.}(2023)]%
        {liu2023lost}
\bibfield{author}{\bibinfo{person}{Nelson~F. Liu}, \bibinfo{person}{Kevin Lin}, \bibinfo{person}{John Hewitt}, \bibinfo{person}{Ashwin Paranjape}, \bibinfo{person}{Michele Bevilacqua}, \bibinfo{person}{Fabio Petroni}, {and} \bibinfo{person}{Percy Liang}.} \bibinfo{year}{2023}\natexlab{}.
\newblock \bibinfo{title}{Lost in the Middle: How Language Models Use Long Contexts}.
\newblock
\newblock
\showeprint[arxiv]{2307.03172}~[cs.CL]


\bibitem[Loshchilov and Hutter(2019)]%
        {loshchilov2019decoupled}
\bibfield{author}{\bibinfo{person}{Ilya Loshchilov} {and} \bibinfo{person}{Frank Hutter}.} \bibinfo{year}{2019}\natexlab{}.
\newblock \bibinfo{title}{Decoupled Weight Decay Regularization}.
\newblock
\newblock
\showeprint[arxiv]{1711.05101}~[cs.LG]


\bibitem[Meta(2023)]%
        {touvron2023llama}
\bibfield{author}{\bibinfo{person}{Meta}.} \bibinfo{year}{2023}\natexlab{}.
\newblock \bibinfo{title}{Llama 2: Open Foundation and Fine-Tuned Chat Models}.
\newblock
\newblock
\showeprint[arxiv]{2307.09288}~[cs.CL]


\bibitem[Nussbaum et~al\mbox{.}(2024)]%
        {nussbaum2024nomic}
\bibfield{author}{\bibinfo{person}{Zach Nussbaum}, \bibinfo{person}{John~X. Morris}, \bibinfo{person}{Brandon Duderstadt}, {and} \bibinfo{person}{Andriy Mulyar}.} \bibinfo{year}{2024}\natexlab{}.
\newblock \bibinfo{title}{Nomic Embed: Training a Reproducible Long Context Text Embedder}.
\newblock
\newblock
\showeprint[arxiv]{2402.01613}~[cs.CL]


\bibitem[OpenAI(2022)]%
        {chatgpt}
\bibfield{author}{\bibinfo{person}{OpenAI}.} \bibinfo{year}{2022}\natexlab{}.
\newblock \showarticletitle{Introducing chatgpt}.
\newblock
\urldef\tempurl%
\url{https://openai.com/blog/chatgpt}
\showURL{%
\tempurl}


\bibitem[OpenAI(2023)]%
        {openai2023gpt4}
\bibfield{author}{\bibinfo{person}{OpenAI}.} \bibinfo{year}{2023}\natexlab{}.
\newblock \bibinfo{title}{GPT-4 Technical Report}.
\newblock
\newblock
\showeprint[arxiv]{2303.08774}~[cs.CL]


\bibitem[Yi(2023)]%
        {Yi}
\bibfield{author}{\bibinfo{person}{Yi}.} \bibinfo{year}{2023}\natexlab{}.
\newblock \bibinfo{title}{A series of large language models trained from scratch by developers at 01-ai}.
\newblock \bibinfo{howpublished}{\url{https://github.com/01-ai/Yi}}.
\newblock


\bibitem[Zaib et~al\mbox{.}(2022)]%
        {10.1007/s10115-022-01744-y}
\bibfield{author}{\bibinfo{person}{Munazza Zaib}, \bibinfo{person}{Wei~Emma Zhang}, \bibinfo{person}{Quan~Z. Sheng}, \bibinfo{person}{Adnan Mahmood}, {and} \bibinfo{person}{Yang Zhang}.} \bibinfo{year}{2022}\natexlab{}.
\newblock \showarticletitle{Conversational question answering: a survey}.
\newblock \bibinfo{journal}{\emph{Knowl. Inf. Syst.}} \bibinfo{volume}{64}, \bibinfo{number}{12} (\bibinfo{date}{dec} \bibinfo{year}{2022}), \bibinfo{pages}{3151–3195}.
\newblock
\showISSN{0219-1377}
\urldef\tempurl%
\url{https://doi.org/10.1007/s10115-022-01744-y}
\showDOI{\tempurl}


\bibitem[Zeng et~al\mbox{.}(2023)]%
        {zeng2022glm}
\bibfield{author}{\bibinfo{person}{Aohan Zeng}, \bibinfo{person}{Xiao Liu}, \bibinfo{person}{Zhengxiao Du}, \bibinfo{person}{Zihan Wang}, \bibinfo{person}{Hanyu Lai}, \bibinfo{person}{Ming Ding}, \bibinfo{person}{Zhuoyi Yang}, \bibinfo{person}{Yifan Xu}, \bibinfo{person}{Wendi Zheng}, \bibinfo{person}{Xiao Xia}, \bibinfo{person}{Weng~Lam Tam}, \bibinfo{person}{Zixuan Ma}, \bibinfo{person}{Yufei Xue}, \bibinfo{person}{Jidong Zhai}, \bibinfo{person}{Wenguang Chen}, \bibinfo{person}{Peng Zhang}, \bibinfo{person}{Yuxiao Dong}, {and} \bibinfo{person}{Jie Tang}.} \bibinfo{year}{2023}\natexlab{}.
\newblock \bibinfo{title}{GLM-130B: An Open Bilingual Pre-trained Model}.
\newblock
\newblock
\showeprint[arxiv]{2210.02414}~[cs.CL]


\end{thebibliography}

\end{document}